\documentclass[
twocolumn,
]{ceurart}

\usepackage{microtype}
\usepackage{graphicx}
\usepackage{subfigure}
\usepackage{multirow}
\usepackage{paralist}
\usepackage{amsmath,amssymb,amsfonts}
\usepackage{booktabs} 
\usepackage{xcolor}

\DeclareMathOperator*{\argmin}{arg\,min}

\begin{document}

\copyrightyear{2020}
\copyrightclause{Copyright for this paper by its authors.
  Use permitted under Creative Commons License Attribution 4.0
  International (CC BY 4.0).}

\conference{Title of the Proceedings: "Proceedings of the CIKM 2020 Workshops" 
Editors of the Proceedings: Stefan Conrad, Ilaria Tiddi}

\title{Now You See Me (CME): Concept-based Model Extraction}

\author[1, 3]{Dmitry Kazhdan}
\ead{dk525@cam.ac.uk}
\address[1]{The University of Cambridge, UK}
\address[2]{The Alan Turing Institute, London, UK}
\address[3]{Denotes equal contribution}

\author[1, 3]{Botty Dimanov}
\ead{btd26@cam.ac.uk}

\author[1]{Mateja Jamnik}
\ead{mateja.jamnik@cl.cam.ac.uk}

\author[1]{Pietro Li\`{o}}
\ead{pietro.lio@cl.cam.ac.uk}

\author[1,2]{Adrian Weller}
\ead{adrian.weller@eng.cam.ac.uk}

\begin{abstract}
Deep Neural Networks (DNNs) have achieved remarkable performance on a range of tasks. A key step to further empowering DNN-based approaches is improving their explainability. In this work we present CME: a concept-based model extraction framework, used for analysing DNN models via concept-based extracted models. Using two case studies (dSprites, and Caltech UCSD Birds), we demonstrate how CME can be used to (i) analyse the concept information learned by a DNN model (ii) analyse how a DNN uses this concept information when predicting output labels (iii) identify key concept information that can further improve DNN predictive performance (for one of the case studies, we showed how model accuracy can be improved by over $14\%$, using only $30\%$ of the available concepts).
\end{abstract}


\maketitle


\section{Introduction} 
\label{introduction}

The black-box nature of Deep Neural Networks (DNNs) hinders their widespread adoption, especially in industries under heavy regulation with high-cost of error \cite{goodman2017european}. As a result, there has recently been a dramatic increase in research on Explainable AI (XAI), focusing on improving explainability of DL systems~\cite{arrieta2020explainable, adadi2018peeking}. 

Currently, the most widely used XAI methods are feature importance methods (also referred to as saliency methods) \cite{Bhatt2019ExplainableML}. For a given data point, these methods provide scores showing the importance of each feature (e.g.,  pixel, patch, or word vector) to the algorithm's decision. Unfortunately, feature importance methods have been shown to be fragile to input perturbations~\cite{kindermans2019reliability,melis2018towards} or model parameter perturbations~\cite{adebayo2018sanity,dimanov2020you}. Human experiments also demonstrate that feature importance explanations do not necessarily increase human understanding, trust, or ability to correct mistakes in a model~\cite{poursabzi2018manipulating,tcav}.

As a consequence, two other types of XAI approaches are receiving increasing attention: \textit{model extraction} approaches, and \textit{concept-based} explanation approaches. Model extraction methods (also referred to as model translation methods) approximate black-box models with simpler models to increase model explainability. Concept-based explanation approaches provide model explanations in terms of human-understandable units, rather than individual features, pixels, or characters (e.g., the concepts of a \textit{wheel} and a \textit{door} are important for the detection of cars)~\cite{tcav,zhou2018interpretable,ghorbani2019towards}. 

\begin{figure} 
    \centering
    
    \subfigure[]{\includegraphics[scale=0.33]{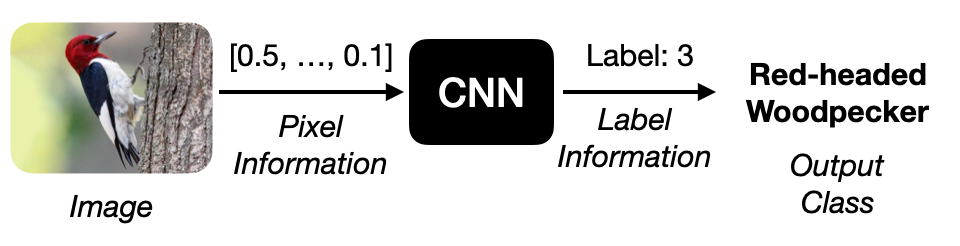}}
    
    \subfigure[]{\includegraphics[scale=0.23]{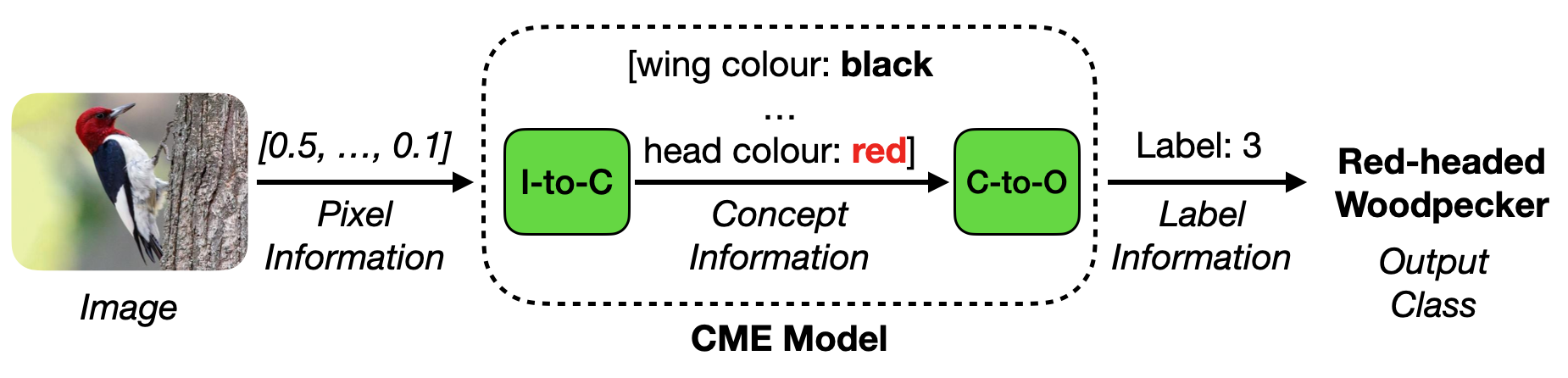}}

    \caption{CME extracted model example. (a) Given an input image, a CNN uses the image's pixel information as input, and returns class information as output (in this case, class label $3$, corresponding to the \textit{Red-headed Woodpecker} class), performing data processing in a non-explainable, black-box fashion. (b) Given an input image, a CME extracted model uses an \textit{Input-to-Concept} function (\textit{I-to-C}) to compute \textit{concept information} from the pixel data (e.g. bird wing color, or head color values). Next, the model uses a \textit{Concept-to-Output} function (\textit{C-to-O}) to compute the output class label from this concept information.}
    \label{fig:cme}
\end{figure}

In this paper we introduce CME\footnote{Pronounced ``See Me.''}: a (C)oncept-based (M)odel (E)xtraction framework\footnote{All relevant code is available at \sep https://github.com/dmitrykazhdan/CME}. Figure~\ref{fig:cme} depicts how CME can be used to analyse DNN models via explainable concept-based extracted models, in order to explain and improve performance of DNNs, as well as to extract useful knowledge from them. Although this example focuses on a CNN model, CME is model-agnostic, and can be applied to \textit{any} DNN architecture. 

In particular, we make the following contributions: 

\begin{compactitem}
    \item  We present the novel CME framework, capable of analysing DNN models via concept-based extracted models
    
    \item  We demonstrate, using two case-studies, how CME can analyse (both quantitatively and qualitatively) the concept information a DNN model has learned, and how this information is represented accross the DNN layers
    
    \item We propose a novel metric for evaluating the quality of concept extraction methods
    
    \item We demonstrate, using two case-studies, how CME can analyse (both quantitatively and qualitatively) how a DNN uses concept information when predicting output labels
    
    \item We demonstrate how CME can identify key concept information that can further improve DNN predictive performance
    
\end{compactitem}

\section{Related Work}

\subsection{Concept-based Explanations} \label{sec:concept-based-expl}

Concept-based explanations have been used in a wide range of different ways, including: inspecting what a model has learned~\cite{ghorbani2019towards, yeh2019concept}, providing class-specific explanations~\cite{kim2017interpretability, tcav}, and discovering causal relations between concepts~\cite{goyal2019explaining}. Similarly to CME, these approaches typically seek to explain model behaviour in terms of high-level concepts, extracting this concept information from a model's latent space.

Importantly, existing concept-based explanation approaches are typically capable of handling binary-valued concepts only, which implies that multi-valued concepts have to be binarised first. For instance, given a concept such as ``shape'', with possible values `square' and `circle', these approaches have to convert ``shape'' into two binary concepts `is\_square', and `is\_circle'. This makes such approaches (i) computationally expensive, since the binarised concept space usually has a high cardinality, (ii) error-prone, since mutual exclusivity of concept values is now not enforced (e.g., a single data point can now have both `is\_square' and `is\_circle' concepts being true). In contrast, our approach is capable of handling multi-valued concepts directly, without binarisation. 

Furthermore, concept-based explanation approaches typically rely on the latent space of a single layer when extracting concept information. DNNs have been shown to perform hierarchical feature extraction, with layers closer to the output utilising higher-level data representations, compared to layers closer to  the  input \cite{hinton2007learning, zhou2014object}. This implies that choosing a single layer imposes an \textit{unnecessary} trade-off between low- and high-level concepts. On the other hand, CME is capable of efficiently combining latent space information from multiple layers, thereby avoiding this constraint.  

Finally, existing methods typically represent concept explanations as a list of concepts, with their relative importance with respect to the classification task. In contrast, our approach describes the functional relationship between concepts and outputs, thereby showing in more detail how the model utilises concept information when making predictions.

\subsection{Concept Bottleneck Models} \label{rel_cbm}

Recent work on concept-based explanations relies on models that use an intermediate concept-based representation when making predictions \cite{koh2020concept, LagDos20}. Work in \cite{koh2020concept} refer to these types of models as \textit{concept bottleneck models} (CBMs). A concept bottleneck model is a model which, given an input, first predicts an intermediate set of human-specified concepts, and then uses only this concept information to predict the output task label. Work in \cite{koh2020concept} proposes a method for turning any DNN into a concept bottleneck model given concept annotations at training time. This is achieved by resizing one of the layers to match the number of concepts provided, and re-training with an added intermediate loss that encourages the neurons in that layer to align component-wise to the provided concepts.

Crucially, CBM approaches provide ways for \textit{generating DNN models}, which are explicitly encouraged to rely on specified concept information. In contrast, our approach is used for \textit{analysing DNN models} (and is much cheaper computationally).

Furthermore, CBM approaches require concept annotations to be available at training time for \textit{all} of the training data, which is often expensive to produce. In contrast, CME can be used with \textit{partially-labelled} datasets in a semi-supervised fashion, as will be described in Section \ref{methodology}.

Finally, CBM approaches require the concepts themselves to be known beforehand. On the other hand, CME can efficiently utilise knowledge contained in pre-trained DNNs, in order to \textit{learn} about which concepts are/aren't required for a given task. Further details on CME/CBM comparison can be found in Appendix \ref{bottleneck_app}.

\subsection{Model Extraction}

Model extraction techniques use rules~\cite{andrews1995survey,zilke2016deepred,chen2017enhancing}, decision trees~\cite{krishnan1999extracting,sato2001rule}, or other more readily explainable models~\cite{kazhdan2020marleme} to approximate complex models, in order to study their behaviour. Provided the approximation quality (referred to as \textit{fidelity}) is high enough, an extracted model can preserve many statistical properties of the original model, while remaining open to interpretation.

However, extracted models generated by existing methods represent their decision-making using the same input representation as the original model, which is typically difficult for the user to understand directly. Instead, our extracted models represent decision-making via human-understandable concepts, making them easier to interpret.




\section{Methodology} 
\label{methodology}

In this section we present our CME approach, describing how it can be used to analyse DNN models using concept-based extracted models.

\subsection{Formulation}

We consider a pre-trained DNN classifier $f: \mathcal{X} \rightarrow \mathcal{Y}$, ($\mathcal{X} \subset \mathbb{R}^n$, $\mathcal{Y} \subset \mathbb{R}^o$), where $f(\mathbf{x})=y$ is mapping an input $\mathbf{x} \in \mathcal{X}$ to an output class $y \in \mathcal{Y}$. For every DNN layer $l$, we denote the function $f^{l} : \mathcal{X} \to \mathcal{H}^l$, ($\mathcal{H}^l \subset \mathbb{R}^m$) as a mapping from the input space $\mathcal{X}$ to the hidden representation space $\mathcal{H}^l$, where $m$ denotes the number of hidden units, and can be different for each layer.

Similarly to \cite{koh2020concept, LagDos20}, we assume the existence of a \textit{concept representation} $\mathcal{C} \subset \mathbb{R}^k$, defining $k$ distinct concepts associated with the input data. $\mathcal{C}$ is defined such that every basis vector in $\mathcal{C}$ spans the space of possible values for one particular concept. We further assume the existence of a function $p^{\star}: \mathcal{X} \rightarrow \mathcal{C}$, where $p^{\star}(\mathbf{x}) = \mathbf{c}$ is mapping an input $\mathbf{x}$ to its concept representation $\mathbf{c}$. Thus, $p^{\star}$ defines the concepts and their values (referred to as the \textit{ground truth concepts}) for every input point. 

\subsection{CME}

In this work, we define a DNN $f$ as being \textit{concept-decomposable}, if it can be well-approximated by a composition of functions $p$ and $q$, such that $f(\mathbf{x})=q(p(\mathbf{x}))$. In this definition, the function $p : \mathcal{X} \to \mathcal{C}$ is an \textit{input-to-concept} function, mapping data-points from their input representation $\mathbf{x} \in \mathcal{X}$ to their concept representation $\mathbf{c} \in \mathcal{C}$. The function $q : \mathcal{C} \to \mathcal{Y}$ is a \textit{concept-to-output} function, mapping data-points in their concept representation $\mathcal{C}$ to output space $\mathcal{Y}$. Thus, when processing an input $\mathbf{x}$, a DNN $f$ can be seen as converting this input into an interpretable concept representation using $p$, and using $q$ to predict the output from this representation. The significance of this decomposition is further discussed in Appendix \ref{bottleneck_app}. 

CME explores whether a given DNN $f$ is concept-decomposable, by attempting to approximate $f$ with an extracted model $\hat{f}: \mathcal{X} \rightarrow \mathcal{Y}$. In this case, $\hat{f}$ is defined as $\hat{f}(\mathbf{x}) = \hat{q}(\hat{p}(\mathbf{x}))$, using input-to-concept $\hat{p}$ and output-to-concept $\hat{q}$ extracted by CME from the original DNN. We describe our approach to extracting $\hat{p}$ and $\hat{q}$ in the remainder of this section.

\subsection{Input-to-Concept ($\hat{p}$)}

When extracting $\hat{p}$ from a pre-trained DNN, we assume we have access to the DNN training data and labels $\{ (\mathbf{x}^{(0)}, y^{(0)}), ..., (\mathbf{x}^{(d)}, y^{(d)}) \}$. Furthermore, we assume partial access to $p^{\star}$, such that a small set of $i$ training points $\{\mathbf{x}^{(0)}, ..., \mathbf{x}^{(i-1)}\}$ have concept labels $\{\mathbf{c}^{(0)}, ..., \mathbf{c}^{(i-1)}\}$ associated with them, while the remaining $u$ points $\{\mathbf{x}^{(i)}, ..., \mathbf{x}^{(i+u)}\}$ do not (in this case $u = d-i$). We refer to these subsets respectively as the \textit{concept labelled dataset} and \textit{concept unlabelled dataset}. Using these datasets, we generate $\hat{p}$ by aggregating concept label predictions across multiple layers of the given DNN model, as described below.

Given a DNN layer $l$ with $m$ hidden units, we compute the layer's representation of the input data $\mathbf{h} = f^l(\mathbf{x})$, obtaining $(\mathbf{h}^{(0)}, ..., \mathbf{h}^{(i+u)})$. Using this data and the concept labels, we construct a semi-supervised dataset, consisting of labelled data $\{(\mathbf{h}^{(0)}, \mathbf{c}^{(0)}), ..., (\mathbf{h}^{(i-1)}, \mathbf{c}^{(i-1)})\}$, and unlabelled data $\{\mathbf{h}^{(i)}, ..., \mathbf{h}^{(i+u)}\}$. 

Next, we rely on Semi-Supervised Multi-Task Learning (SSMTL) \cite{liu2008semi}, in order to extract a function $g^{l} : \mathcal{H}^l \to \mathcal{C}$, which predicts concept labels from layer $l$'s hidden space. In this work, we treat each concept as a separate, independent task. Hence, $g^{l}(\mathbf{h})$ is decomposed into $k$ separate tasks (one per concept), and is defined as $g^{l}(\mathbf{h}) = (g^{l}_{1}(\mathbf{h}), ...,  g^{l}_{k}(\mathbf{h}))$ where each $g^{l}_{i}(\mathbf{h})$ ($i \in \{ 1..k \} $) predicts the value of concept $i$ from $\mathbf{h}$.

Repeating this process for all model layers $L$, we obtain a set of functions $G = \{g^{l}_{i} \ | \ l \in \{1..L\} \ \land i \in \{1..k\} \}$. For every concept $i$, we define the ``best'' layer $l^{i}$ for predicting that concept as shown in \eqref{eq:top_g_eqn}:  

\begin{equation} 
\label{eq:top_g_eqn}
l^{i} = \argmin_{l \in L}\ell(g^l_{i}, i)
\end{equation}

Here, $\ell$ is a loss function (in this case the error rate), computing the predictive loss of function $g^l_i$ with respect to a concept $i$. Finally, we define $\hat{p}$ as shown in \eqref{eq:phat}:

\begin{equation} 
\label{eq:phat}
\hat{p}(\mathbf{x}) = (g^{l^{1}}_{1}\circ f^{l^{1}}(\mathbf{x}), ..., g^{l^{k}}_{k}\circ f^{l^{k}}(\mathbf{x}))
\end{equation}

Thus, given an input $\mathbf{x}$, the value computed by $\hat{p}(\mathbf{x})$ for every concept $i \in \{1..k\}$ is equal to the value computed by $g^{l^{i}}_{i}$ from that input's representation in layer $l^{i}$. Overall, $\hat{p}$ encapsulates concept information contained in a given DNN model, and can be used to analyse how this information is represented, as well as to predict concept values for new inputs.

\subsection{Concept-to-Label ($\hat{q}$)}

We setup extraction of $\hat{q}$ as a classification problem, in which we train $\hat{q}$ to predict output labels $y$ from concept labels $\mathbf{c}$ predicted by $\hat{p}$. We use $\hat{p}$ to generate concept labels for all training data points, obtaining a set of concept labels $\{\mathbf{c}^{(0)}, ..., \mathbf{c}^{(i+u)}\}$. Next, we produce a labelled dataset, consisting of concept labels and corresponding DNN output labels $\{(\mathbf{c}^{(0)}, y^{(0)}), ..., (\mathbf{c}^{(i+u)}, y^{(i+u)})\}$, and use it to train $\hat{q}$ in a supervised manner. We experimented with using Decision Trees (DTs), and Logistic Regression (LR) models for representing $\hat{q}$, as will be discussed in Section \ref{results}. Overall, $\hat{q}$ can be used to analyse how a DNN uses concept information when making predictions.


\section{Experimental Setup} \label{exp_setup}

We evaluated CME using two datasets: dSprites~\cite{dsprites17}, and Caltech-UCSD birds~\cite{wah2011caltech}. All relevant code is publicly available at\footnote{https://github.com/dmitrykazhdan/CME}.

\subsection{dSprites}
dSprites is a well-established dataset used for evaluating unsupervised latent factor disentanglement approaches. dSprites consists of 2D $64 \times 64$ pixel black-and-white shape images, procedurally generated from all possible combinations of 6 ground truth independent concepts (color, shape, scale, rotation, x and y position). Further details can be found in Appendix \ref{dsprites_a}, and the official dSprites repository. \footnote{https://github.com/deepmind/dsprites-dataset/}

\subsubsection{Classification Tasks} \label{exp_dspr_cls}
We define $2$ classification tasks, used to evaluate our framework:

\begin{compactitem}
    \item \textbf{Task 1}: This task consists of determining the shape concept value from an input image. For every image sample, we define its task label as the shape concept label of that sample.
    
    \item\textbf{Task 2}: This task consists of discriminating between all possible \textit{shape} and \textit{scale} concept value combinations. We assign a distinct identifier to each possible combination of the shape and scale concept labels. For every image sample, we define its task label as the identifier corresponding to this sample's shape and scale concept values.
    
\end{compactitem}

Overall, Task 1 explores a scenario in which a DNN has to learn to recognise a specific concept from an input image. Task 2 explores a relatively more complex scenario, in which a DNN has to learn to recognise combinations of concepts from an input image.

\subsubsection{Model}
We trained a Convolutional Neural Network (CNN) model \cite{lecun1990handwritten} for each task. Both models had the same architecture, consisting of 3 convolutional layers, 2 dense layers with ReLUs, 50\% dropout \cite{dropout} and a softmax output layer. The models were trained using categorical cross-entropy loss, and achieved $100.0 \pm 0.0 \%$ classification accuracies on their respective held-out test sets. We refer to these models as the \textit{Task 1 model} and the \textit{Task 2 model} in the rest of this work.

\subsubsection{Ground-truth Concept Information}
Importantly, the task and dataset definitions described in this section imply that we know precisely which concepts the models had to learn, in order to achieve $100.0 \pm 0.0 \%$ task performances (\textit{shape} for Task 1, and \textit{shape} and \textit{scale} for Task 2). We refer to this as the \textit{ground truth} concept information learned by these models.

\subsection{Caltech-UCSD Birds (CUB)}
For our second dataset, we used Caltech-UCSD Birds 200 2011 (CUB). This dataset consists of 11,788 images of 200 bird species with every image annotated using 312 binary concept labels (e.g. beak and wing colour, shape, and pattern). We relied on concept pre-processing steps defined in \cite{koh2020concept} (used for de-noising concept annotations, and filtering out outlier concepts), which produces a refined set of $k=112$ binary concept labels for every image sample.

\subsubsection{Classification Task.} 
We relied on the standard CUB classification task, which consists of predicting the bird species from an input image.

\subsubsection{Model} \label{cub_model}
We used the Inception-v3 architecture \cite{szegedy2016rethinking}, pretrained on ImageNet \cite{krizhevsky2012imagenet} (except for the fully-connected layers) and fine-tuned end-to-end on the CUB dataset, following the preprocessing practices described in \cite{cui2018large}. The model achieved $82.7 \pm 0.4 \%$ classification accuracy on a held-out test set. We refer to this model as the \textit{CUB model} in the rest of this work.

\subsubsection{Ground-truth Concept Information} \label{cub_gt_cinf}
Unlike dSprites, the CUB dataset does not explicitly define how the available concepts relate to the output task. Thus, we \textit{do not} have access to the ground truth concept information learned by the CUB model.

\subsection{Benchmarks}

We compare performance of our CME approach to two other benchmarks, described in the remainder of this section.

\subsubsection{Net2Vec}

We rely on work in \cite{fong2018net2vec} for defining benchmark $\hat{p}$ functions for the three tasks. Work in \cite{fong2018net2vec} attempts to predict presence/absence of concepts from spatially-averaged hidden layer activations of convolutional layers of a CNN model. Given a binary concept $c$, this approach trains a logistic regressor, predicting the presence/absence of this concept in an input image from the latent representation of a given CNN layer. In case of multi-valued concepts, the concept space has to be binarised, as discussed in Section \ref{rel_cbm}. In this case, the binarised concept value with the highest likelihood is returned.

Unlike CME, \cite{fong2018net2vec} does not provide a way of selecting the convolutional layer to use for concept extraction. We consider the best-case scenario by selecting, for all tasks, the convolutional layers yielding the best concept extraction performance. For all tasks, these layers were convolutional layers closest to the output (the 3rd conv. layer in case of dSprites tasks, and the final inception block output layer in case of the CUB task).

\subsubsection{CBM}

As discussed in Section \ref{cub_gt_cinf}, we do not have access to ground truth concept information learned by the CUB model. Instead, we rely on the pre-trained \textit{sequential bottleneck model} defined in \cite{koh2020concept} (referred to as CBM in the rest of this work). CBM is a bottleneck model, obtained by resizing one of the layers of the CUB model to match the number of concepts provided (we refer to this as the \textit{bottleneck layer}), and training the model in two steps. First, the sub-model consisting of the layers between the input layer and the bottleneck layer (inclusive) is trained to predict concept values from input data. Next, the submodel consisting of the layers between the layer following the bottleneck layer and the output layer is trained to predict task labels from the concept values predicted by the first submodel. Hence, this bottleneck model is \textit{guaranteed} to solely rely on concept information that is learnable from the data, when making task label predictions. Thus, this benchmark serves as an \textit{upper bound} for the concept information learnable from the dataset, and for the task performance achievable using this information. Importantly, CBM does not attempt to approximate/analyse the CUB model, but instead attempts to solve the same classification task using concept information only. 

We use the first CBM submodel as a $\hat{p}$ benchmark, representing the upper bound of concept information learnable from the data. We use the second submodel as a $\hat{q}$ benchmark, representing the upper bound of task performance achievable from predicted concept information only. Finally, we use the entire model as an $\hat{f}$ benchmark. We make use of the saved trained model from \cite{koh2020concept}, available in their official repository\footnote{https://github.com/yewsiang/ConceptBottleneck}.


\section{Results} \label{results}

We present the results obtained by evaluating our approach using the two case studies described  above. 


We obtain the concept labelled dataset by returning the ground-truth concept values for a random set of samples in the model training data. For dSprites, we found that a concept labelled dataset of a $100$ samples or more worked well in practice for both tasks. Thus, we fix the size of the concept labelled dataset to $100$ in all of the dSprites experiments. For CUB, we found that a concept labelled dataset containing $15$ or more samples per class worked well in practice. Thus, we fix the size of the concept labelled dataset to $15$ samples per class in all of the CUB experiments. In the future, we intend to explore the variation of model extraction performance with the size of the concept labelled dataset in more detail.

\subsection{Concept Prediction Performance} \label{phat_perf}

First, we evaluate the quality of $\hat{p}$ functions produced by CME, Net2Vec, and CBM. For both dSprites tasks, we relied on the \textit{Label Spreading} semi-supervised model \cite{Zhou04learningwith}, provided in scikit-learn \cite{scikit-learn}, when learning the $g^{l}_{i}$ functions for CME. For CUB, we used logistic regression functions instead, as they gave better performance.

\subsubsection{dSprites}

\begin{figure} 
\centering
\subfigure[Task 1]{\includegraphics[width=0.48\linewidth]{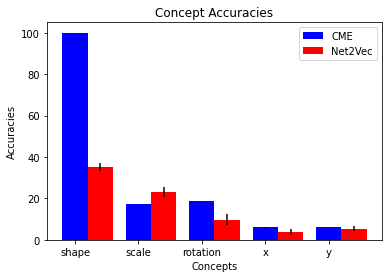}}
\subfigure[Task 2]{\includegraphics[width=0.48\linewidth]{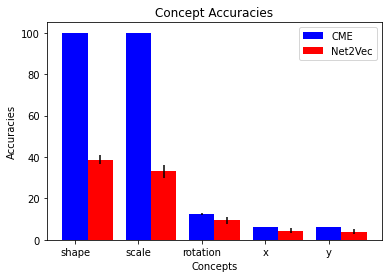}}

\caption{Predictive accuracy of CME and Net2Vec $\hat{p}$ functions for all concepts}
\label{phat_fig}
\end{figure}

Figure \ref{phat_fig} shows predictive performance of the $\hat{p}$ functions on all concepts for the two dSprites tasks (averaged over 5 runs). As discussed in Section \ref{exp_dspr_cls}, we have access to the ground truth concept information learned by these models (\textit{shape} concept information for Task 1, and \textit{shape} and \textit{scale} concept information for Task 2). For both tasks, $\hat{p}$ functions extracted by CME successfully achieved high predictive accuracy on concepts relevant to the tasks, whilst achieving a low performance on concepts irrelevant to the tasks. Thus, CME was able to successfully extract the concept information contained in the task models. For both tasks, $\hat{p}$ functions extracted by Net2Vec achieved a much lower performance on the relevant concepts.

\subsubsection{CUB}

As discussed in Section \ref{cub_gt_cinf}, the CUB dataset does not explicitly define how the concepts relate to the output task labels. Thus, we do not know how relevant/important different concepts are, with respect to task label prediction. In this section, we make the conservative assumption that all concepts are relevant, when evaluating $\hat{p}$ functions, and explore relative concept importance in more detail in Section \ref{intervention_perf}. 

Firstly, we relied on the `average-per-concept' metrics introduced in \cite{koh2020concept} when evaluating the $\hat{p}$ function performances, by computing their $F1$ predictive scores for each concept, and then averaging over all concepts. We obtained $F1$ scores of 92 $\pm$ 0.5\%, 86.3 $\pm$ 2.0\%, and 85.9 $\pm$ 2.3\% for CBM, CME, and Net2Vec $\hat{p}$ functions, respectively (averaged over 5 runs).

Importantly, we argue that in case of a large number of concepts, it is crucial to measure how concept mispredictions are distributed accross the test samples. For instance, consider a dSprites Task 2 $\hat{p}$ function that achieves $90\%$ predictive accuracy on both \textit{shape} and \textit{scale} concepts. The average predictive accuracy on relevant concepts achieved by this $\hat{p}$ will therefore be $90\%$. However, if the two concepts are mis-predicted for strictly different samples (i.e. none of the samples have both \textit{shape} and \textit{scale} predicted incorrectly at the same time), this means that $20\%$ of the test samples will have one relevant concept predicted incorrectly. Given that both concepts need to be predicted correctly when using them for task label prediction, this implies that consequent task label prediction will not be able to achieve over $80\%$ task label accuracy. This effect becomes even more pronounced in case of a larger number of relevant concepts.

Consequently, we defined a novel cumulative misprediction error metric, which we refer to as the `mis-prediction-overlap' (MPO) metric. Given a test set $T = \{(\mathbf{x}^{(0)}, \mathbf{c}^{(0)}),$ $ ..., (\mathbf{x}^{(n)}, \mathbf{c}^{(n)})\}$ consisting of $n+1$ input samples $\mathbf{x}$ with corresponding concept labels $\mathbf{c}$, and a prediction set  $P = \{(\mathbf{\hat{c}}^{(0)}), ..., \mathbf{\hat{c}}^{(n)}\}$, $MPO$ computes the fraction of samples in the test set, that have \textit{at least} $m$ relevant concepts predicted incorrectly, as shown in Equation \ref{eq:mpo} (where $\mathbb{I}(.)$ denotes the indicator function):

\begin{equation} 
\label{eq:mpo}
MPO(T, P, m) = \frac{1}{n} \sum\limits_{i=0}^n \mathbb{I} (err(\mathbf{c}_{i}, \mathbf{\hat{c}}_{i}) >= m)
\end{equation}

Here, $err$ can be used to specify which concepts to measure the mis-prediction error on (i.e. in case some of the provided concepts are irrelevant). Under our assumption of all concepts being relevant, we defined $err$ as shown in Equation \ref{eq:error}:

\begin{equation} 
\label{eq:error}
err(\mathbf{c}_{i}, \mathbf{\hat{c}}_{i}) = \sum\limits_{j=0}^k \mathbb{I}(c_{i, j} \neq \hat{c}_{i, j})
\end{equation}

\begin{figure} 
    \centering
    \includegraphics[scale=0.3]{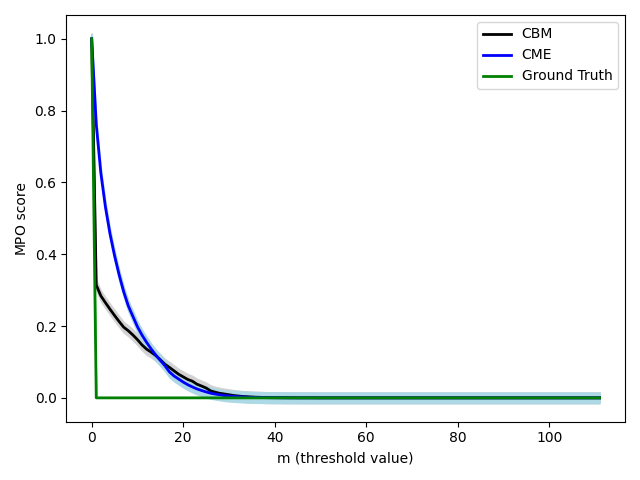}
    \caption{Performances of $\hat{p}$ functions, evaluated using the $MPO$ metric. The green line plots the case for perfect prediction, when the predicted concepts are equivalent to the ground truth concepts (i.e. the $p^{\star}$ performance), in which case $MPO = 1$ for $m=0$, and $MPO=0$ otherwise. Net2Vec obtained values within $1\%$ deviation from the corresponding CME values for all $m$, and is therefore omitted here for simplicity}
    \label{fig:p_hat_cub}
\end{figure}

Using a held-out test set, we plot the $MPO$ metric values for $m \in [0, ..., 112]$, as shown in Figure \ref{fig:p_hat_cub} (averaged over 5 runs). Importantly, $\hat{p}$ function performances can be evaluated by observing their $MPO$ scores for different values of $m$. A larger $MPO$ score implies a bigger proportion of samples had at least $m$ relevant concept predicted incorrectly.

Overall, CME performed almost identically to Net2Vec, and worse than $CBM$ according to the $MPO$ metric. Similar performance to Net2Vec is likely caused by (i) concepts being binary (requiring no binarisation) (ii) the Inception-v3 model having a relatively large number of convolutional layers, implying that the final convolutional layer likely learned higher-level features, relevant to concept prediction. 

Importantly, $MPO$ showed that both CBM and CME $\hat{p}$ functions had a significant proportion of test samples with incorrectly-predicted relevant concepts (e.g. CME had an MPO score of $0.25$ at $m=4$, implying that 25\% of all test samples have at least 4 concepts predicted incorrectly). In practice, these mispredictions can have a significant impact on consequent task label predictive performance, as will be further explored in the next section.

\subsection{Task Performance} \label{qhat_perf}

In this section, we evaluate the fidelity and performance of the extracted $\hat{f}$ models. For all CME and Net2Vec $\hat{p}$ functions evaluated in the previous section, we trained output-to-concept functions $\hat{q}$, predicting class labels from the $\hat{p}$ concept predictions. Next, for every $\hat{p}$, we defined its corresponding $\hat{f}$ as discussed in Section \ref{methodology}, via a composition of $\hat{p}$ and its associated $\hat{q}$. For every $\hat{f}$, we evaluated its fidelity and its task performance, using a held-out sample test set. Table~\ref{tbl:fidelity_tbl} shows the fidelity of extracted models, and Table~\ref{tbl:perf_tbl} shows the task performance for these models (averaged over 5 runs). The original Task 1, Task 2, and CUB models achieved task performances of 100$\pm$0\%, 100$\pm$0\%, and 82.7$\pm$0.4\%, respectively, as described in Section \ref{exp_setup}.

\begin{table}[pos=htbp]
\caption{Fidelity of extracted $\hat{f}$ models}
\begin{center}
\begin{tabular}{cccc}
\hline
\textbf{}           &  \textbf{CME}         & \textbf{CBM}          & \textbf{Net2Vec}  \\ \hline


\textbf{Task 1}     &  100.0$\pm$0.0\%    &  --                     & 24.5$\pm$3.6\%        \\ \hline
\textbf{Task 2}     &  99.3$\pm$0.5\%     &  --                     & 38.3$\pm$4.0\%        \\ \hline
\textbf{CUB}        &  74.42$\pm$3.1\%    &  77.5$\pm$0.2\%         & 73.8$\pm$2.8\%     \\ \hline

\end{tabular}
\end{center}
\label{tbl:fidelity_tbl}
\end{table}

\begin{table}[pos=htbp]
\caption{Task performance of extracted $\hat{f}$ models}
\begin{center}
\begin{tabular}{ccccc}
\hline
\textbf{}               & \textbf{CME}      &  \textbf{CBM}     & \textbf{Net2Vec} \\ \hline


\textbf{Task 1}     & 100.0$\pm$0\%     & --                & 24.5$\pm$3.6\%   \\ \hline
\textbf{Task 2}     & 99.3$\pm$0.5\%    &  --               & 38.3$\pm$4.0\%   \\ \hline
\textbf{CUB}        & 70.8$\pm$1.8\%    & 75.7$\pm$0.6\%    & 69.8$\pm$1.5\%   \\\hline

\end{tabular}
\end{center}
\label{tbl:perf_tbl}
\end{table}

For both dSprites tasks, CME $\hat{f}$ models achieved high (99\%+) fidelity and task performance scores, indicating that CME successfully approximated the original dSprites models. Furthermore, these scores were considerably higher than those produced by the Net2Vec $\hat{f}$ models.

For the CUB task, both CME and Net2Vec $\hat{f}$ models achieved relatively lower fidelity and task performance scores (in this case, performance of CME was very similar to that of Net2Vec). Crucially, the CBM model \textit{also} achieved relatively low fidelity and accuracy scores (as anticipated from our $MPO$ metric analysis). This implies that concept information learnable from the data is insufficient for achieving high task accuracy. Hence the relatively high CUB model accuracy has to be caused by the CUB model relying on other non-concept information. Thus, the low fidelity of CME and Net2Vec is a consequence of the CUB model being \textit{non-concept-decomposable}, implying that it's behaviour cannot be explained by the desired concepts. The next section discusses possible approaches to fixing this issue.

\subsection{Intervening} \label{intervention_perf}

In the previous section, we demonstrated how CME can be used to identify whether a model relies on desired concepts during decision-making. In this section, we demonstrate how CME can be used to suggest \textit{model improvements}, aligning model behaviour with the desired concepts. 

We trained a logistic regression $\hat{q}$ model predicting task labels from ground-truth concept labels for the CUB task, obtaining an accuracy score of 96.4 $\pm$ 0.5\% on a held-out test set (averaged over 5 runs). Using this model's coefficient magnitudes as a measure of concept importance, we discovered that the $32$ most important concepts identified this way were sufficient for achieving over 96\% task accuracy using logistic regression.

\begin{figure} 
\centering
\includegraphics[scale=0.3]{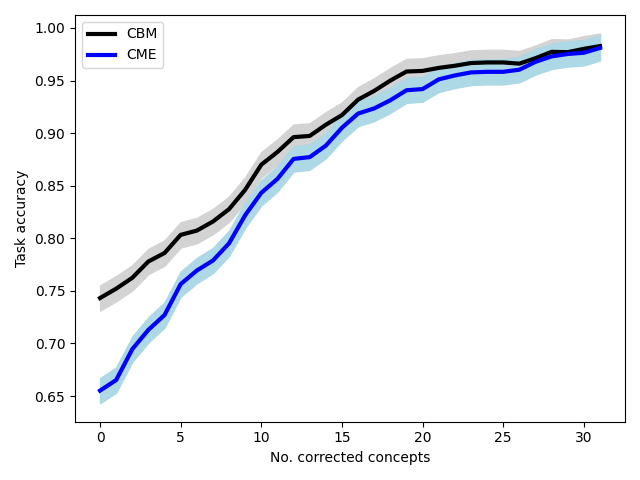} 
\caption{The task accuracy of $\hat{q}$ functions, trained on concepts predicted by $\hat{p}$ functions, with top \# \textit{No. corrected concepts} set to their ground truth values. Performance of Net2Vec was very similar to that of CME, and is thus omitted here for simplicity.}
\label{fig:cub_intervention}
\end{figure}

Using this reduced concept set, we inspected how our CUB $\hat{q}$ function performances would change, if their corresponding $\hat{p}$ functions extracted these concepts perfectly. This was achieved by taking the $\hat{p}$ concept predictions of these concepts on the test and training sets, setting the values of the top $i$ most important concepts to their ground truth values, training logistic regression $\hat{q}$ functions on these modified training sets, and measuring their accuracies on the modified test sets (this approach is referred to as \textit{concept intervention} in the rest of this work). The results are shown in Figure \ref{fig:cub_intervention}, with $i$ ranging from $0$ to $32$.

These results demonstrate that concept information from only $32$ concepts is sufficient for achieving over $96\%$ task performance. Thus, predictive performance of the CUB model can be significantly improved (up to $14\%$) by ensuring that the model is able to learn and use this concept information. Crucially, these results show that CME concept intervention also significantly improves CBM model performance, indicating that the necessary concept information is \textit{not learnable from the data}. Hence, undesired CUB model behaviour is likely arising due to data properties (e.g. the data not being representative with respect to key concepts), not model properties (e.g. architecture, or training regime).

Overall, we demonstrated how CME can be used to identify the key concept information that can be used to improve performance of DNN models, and ensure that they are closer aligned with the desired concept-based behaviour. Furthermore, we demonstrated how CME can be used to identify whether undesired model behaviour is caused by model properties, or data properties.

\subsection{Explainability} \label{res_expl}

By studying CME-extracted $\hat{p}$ and $\hat{q}$ functions separately, we can gain additional insights into what concept information the original model learned and how this concept information is used to make predictions. We give examples of how these sub-models can be inspected in the remainder of this section.

\subsubsection{Input-to-Concept ($\hat{p}$)} \label{sec:eval-input-to-concept}

\begin{figure} 
\centering

\includegraphics[width=\linewidth]{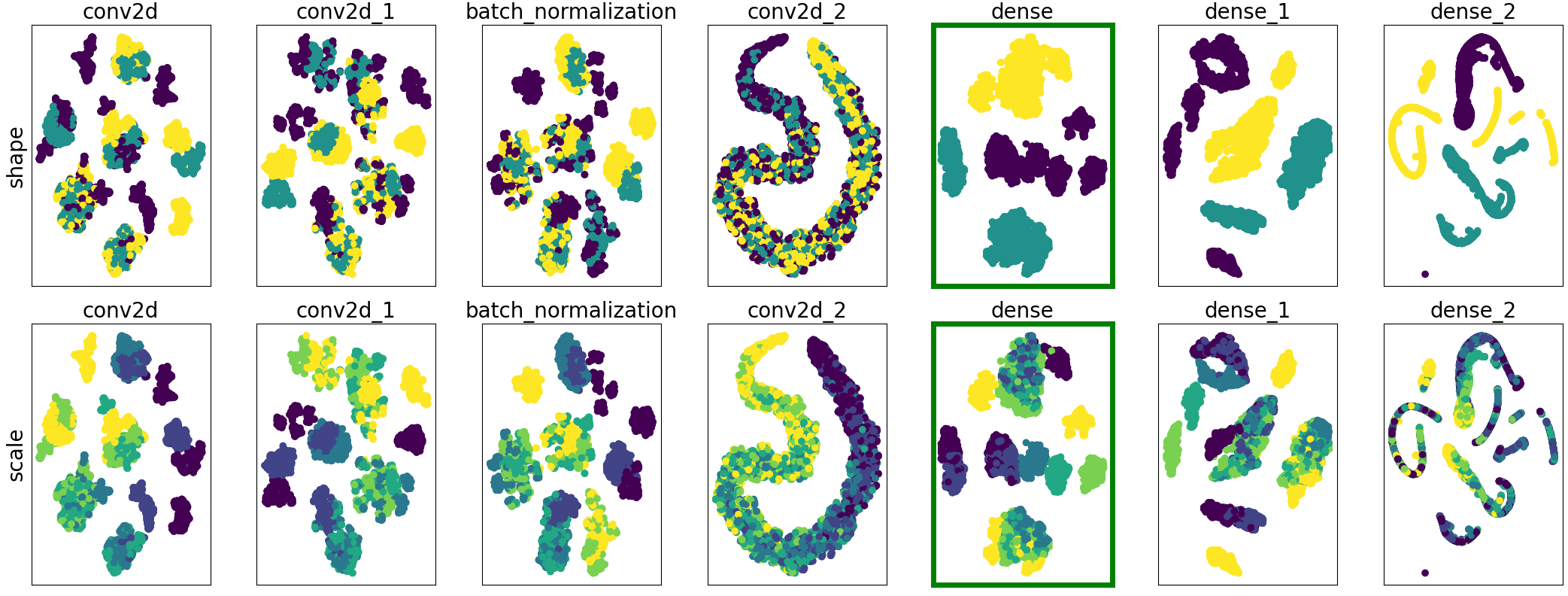} 
\caption{t-SNE plots for the relevant Task 2 concepts. Each row corresponds to a different concept, and each column corresponds to a different layer of the Task 2 model. Each plot is colored with respect to the concept's values. For every concept row, the subplot with a green border indicates the layer CME selected for predicting the value of that concept.}
\label{fig:tsne}
\end{figure}

CME extraction of $\hat{p}$ functions from a DNN model is highly complementary to existing approaches on latent space analysis. For example, Figure~\ref{fig:tsne} shows a t-SNE~\cite{maaten2008visualizing} 2D projected plot of every layer's hidden space of the dSprites Task 2 model, highlighting different concept values of the two relevant concepts, as well as the layers used by CME to predict them. Figure \ref{fig:tsne} demonstrates several important ways in which CME concept extraction can be combined with existing latent space analysis approaches, which will be discussed in the remainder of this section. Further examples are given in Appendix \ref{i_to_c_app}.

\paragraph{Manifold Types} Using ground-truth concept information and hidden space visualisation, it is possible to inspect the nature of latent space manifolds, with respect to specific concepts. Firstly, this inspection allows to build an intuition of how concept information is represented in a particular latent space. Secondly, it is possible to use this information when selecting the types of $\hat{p}$ functions to use during concept extraction. For instance, some manifolds consist of ``blobs'' encoding distinct concept values (e.g. row \verb|shape|, columns \verb|dense|, \verb|dense_1|), suggesting that the latent space is clustered with respect to a concept's values.


\paragraph{Variation Across Layers} Using ground-truth concept information and hidden space visualisation, it is also possible to inspect how concept information representation varies across layers of a DNN model. Firstly, this inspection allows to build an intuition of how concept-related information is transformed by the DNN. Secondly, it is possible to use this information to identify the `best' layers to extract concept information from. For instance, both rows \verb|shape| and \verb|scale| illustrate that the manifolds of higher layers become more unimodal (separating concept values) with respect to the relevant concepts. Importantly, this analysis, together with the definition of $\hat{p}$ allows using \textit{different} layers for extracting different concepts.

Overall, we argue that CME concept extraction can be well-integrated with existing latent space analysis approaches, in order to study which concept information is learned by a DNN, and how this information is represented across DNN layers. This type of inspection can have numerous applications, including: (i) inspecting which concepts a model has learned, and verifying whether it has learned the desired concepts (useful for \textit{model explanations} and \textit{model verification}), (ii) inspecting how concept information is represented across different layers (useful for fine-grained \textit{model analysis}), (iii) extracting concept predictions from a DNN (useful for \textit{knowledge extraction}). Further examples and analysis of extracted $\hat{p}$ functions can be found in Appendix \ref{i_to_c_app}.

\subsubsection{Concept-to-Output ($\hat{q}$)}

\begin{figure} 
    \centering
    \includegraphics[scale=0.16]{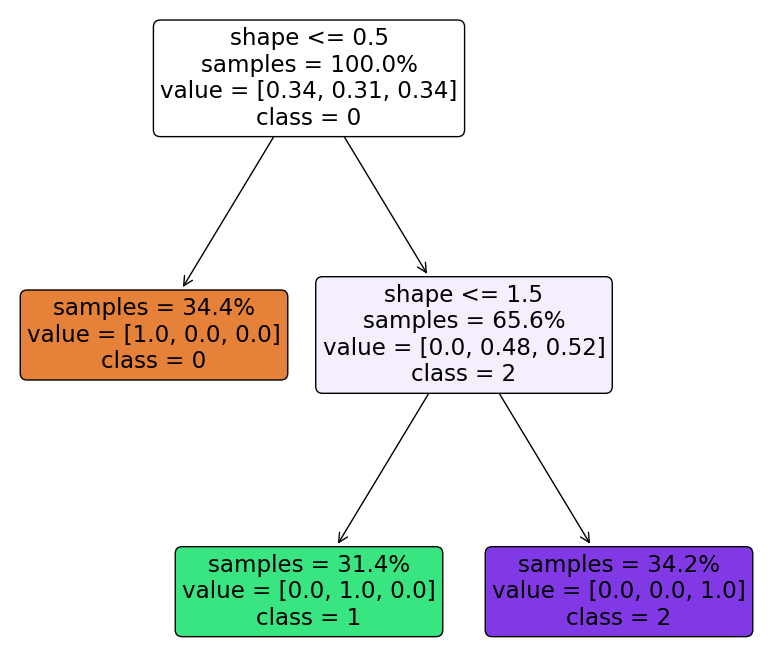}
    \caption{Visualisation of a decision tree $\hat{q}$ extracted from the Task 1 model. The model has correctly learned to differentiate between classes based on the shape concept values.}
    \label{fig:output-to-concept}
\end{figure}

$\hat{q}$ functions encapsulate how a DNN uses concept information when making predictions. Hence, these functions can be inspected directly, in order to analyse model behaviour \textit{represented in terms of concepts}. An example is given in Figure~\ref{fig:output-to-concept}, in which we plot the decision tree $\hat{q}$ function extracted by CME from the Task 1 model. Further examples are given in Appendix \ref{c_t_o_app}.

Overall, inspection of $\hat{q}$ functions can be used for (i) verifying that a DNN uses concept information correctly during decision-making, and that it's high-level behaviour is consistent with user expectations (\textit{model verification}), (ii) identifying specific concepts or concept interactions (if any) causing incorrect behaviour (\textit{model debugging}), (iii) extracting new knowledge about how concept information can be used for solving a particular task (\textit{knowledge extraction}). Further examples and analysis of extracted $\hat{q}$ functions can be found in Appendix \ref{c_t_o_app}.


\section{Conclusions}

We present CME: a concept-based model extraction framework, used for analysing DNN models via concept-based extracted models. Using two case-studies, we demonstrate how CME can be used to (i) analyse concept information learned by DNN models (ii) analyse how DNNs use concept information when making predictions (iii) identifying key concept information that can further improve DNN predictive performance. CME is a model-agnostic, general-purpose framework, which can be combined with a wide variety of different DNN models and corresponding tasks.

In this work, we assume a fixed set of concept labels available to CME before model extraction begins (i.e. the concept-labelled dataset). In the future, we intend to explore active-learning based approaches to obtaining maximally-informative concept labels in an interactive fashion. Consequently, these approaches will improve extracted model fidelity by retrieving the most informative concept labels, and reduce manual concept labelling effort.

Given the rapidly-increasing interest in concept-based explanations of DNN models, we believe our approach can play an important role in providing granular concept-based analyses of DNN models.

\section*{Acknowledgements}
AW acknowledges support from the David MacKay Newton research fellowship at Darwin College, The Alan Turing Institute under EPSRC grant EP/N510129/1 \& TU/B/000074, and the Leverhulme Trust via the Leverhulme Centre for the Future of Intelligence (CFI). BD acknowledges support from EPSRC Award \#1778323. DK acknowledges support from EPSRC ICASE scholarship and GSK. DK and BD acknowledge the experience at Tenyks as fundamental to developing this research idea. 

\bibliography{references}

\begin{thebibliography}{37}
\expandafter\ifx\csname natexlab\endcsname\relax\def\natexlab#1{#1}\fi
\providecommand{\url}[1]{\texttt{#1}}
\providecommand{\href}[2]{#2}
\providecommand{\path}[1]{#1}
\providecommand{\DOIprefix}{doi:}
\providecommand{\ArXivprefix}{arXiv:}
\providecommand{\URLprefix}{URL: }
\providecommand{\Pubmedprefix}{pmid:}
\providecommand{\doi}[1]{\href{http://dx.doi.org/#1}{\path{#1}}}
\providecommand{\Pubmed}[1]{\href{pmid:#1}{\path{#1}}}
\providecommand{\bibinfo}[2]{#2}
\ifx\xfnm\relax \def\xfnm[#1]{\unskip,\space#1}\fi
\bibitem[{Goodman and Flaxman(2017)}]{goodman2017european}
\bibinfo{author}{B.~Goodman}, \bibinfo{author}{S.~Flaxman},
\newblock \bibinfo{title}{European union regulations on algorithmic
  decision-making and a “right to explanation”},
\newblock \bibinfo{journal}{AI magazine} \bibinfo{volume}{38}
  (\bibinfo{year}{2017}) \bibinfo{pages}{50--57}.
\bibitem[{Arrieta et~al.(2020)Arrieta, D{\'\i}az-Rodr{\'\i}guez, Del~Ser,
  Bennetot, Tabik, Barbado, Garc{\'\i}a, Gil-L{\'o}pez, Molina, Benjamins
  et~al.}]{arrieta2020explainable}
\bibinfo{author}{A.~B. Arrieta}, \bibinfo{author}{N.~D{\'\i}az-Rodr{\'\i}guez},
  \bibinfo{author}{J.~Del~Ser}, \bibinfo{author}{A.~Bennetot},
  \bibinfo{author}{S.~Tabik}, \bibinfo{author}{A.~Barbado},
  \bibinfo{author}{S.~Garc{\'\i}a}, \bibinfo{author}{S.~Gil-L{\'o}pez},
  \bibinfo{author}{D.~Molina}, \bibinfo{author}{R.~Benjamins}, et~al.,
\newblock \bibinfo{title}{Explainable artificial intelligence (xai): Concepts,
  taxonomies, opportunities and challenges toward responsible ai},
\newblock \bibinfo{journal}{Information Fusion} \bibinfo{volume}{58}
  (\bibinfo{year}{2020}).
\bibitem[{Adadi and Berrada(2018)}]{adadi2018peeking}
\bibinfo{author}{A.~Adadi}, \bibinfo{author}{M.~Berrada},
\newblock \bibinfo{title}{Peeking inside the black-box: A survey on explainable
  artificial intelligence (xai)},
\newblock \bibinfo{journal}{IEEE Access} \bibinfo{volume}{6}
  (\bibinfo{year}{2018}).
\bibitem[{Bhatt et~al.(2020)Bhatt, Xiang, Sharma, Weller, Taly, Jia, Ghosh,
  Puri, Moura, and Eckersley}]{Bhatt2019ExplainableML}
\bibinfo{author}{U.~Bhatt}, \bibinfo{author}{A.~Xiang},
  \bibinfo{author}{S.~Sharma}, \bibinfo{author}{A.~Weller},
  \bibinfo{author}{A.~Taly}, \bibinfo{author}{Y.~Jia},
  \bibinfo{author}{J.~Ghosh}, \bibinfo{author}{R.~Puri}, \bibinfo{author}{J.~M.
  Moura}, \bibinfo{author}{P.~Eckersley},
\newblock \bibinfo{title}{Explainable machine learning in deployment},
\newblock in: \bibinfo{booktitle}{Proceedings of the 2020 Conference on
  Fairness, Accountability, and Transparency}, \bibinfo{year}{2020}, pp.
  \bibinfo{pages}{648--657}.
\bibitem[{Kindermans et~al.(2019)Kindermans, Hooker, Adebayo, Alber,
  Sch{\"u}tt, D{\"a}hne, Erhan, and Kim}]{kindermans2019reliability}
\bibinfo{author}{P.-J. Kindermans}, \bibinfo{author}{S.~Hooker},
  \bibinfo{author}{J.~Adebayo}, \bibinfo{author}{M.~Alber},
  \bibinfo{author}{K.~T. Sch{\"u}tt}, \bibinfo{author}{S.~D{\"a}hne},
  \bibinfo{author}{D.~Erhan}, \bibinfo{author}{B.~Kim},
\newblock \bibinfo{title}{The (un) reliability of saliency methods},
\newblock in: \bibinfo{booktitle}{Explainable AI: Interpreting, Explaining and
  Visualizing Deep Learning}, \bibinfo{publisher}{Springer},
  \bibinfo{year}{2019}, pp. \bibinfo{pages}{267--280}.
\bibitem[{Melis and Jaakkola(2018)}]{melis2018towards}
\bibinfo{author}{D.~A. Melis}, \bibinfo{author}{T.~Jaakkola},
\newblock \bibinfo{title}{Towards robust interpretability with self-explaining
  neural networks},
\newblock in: \bibinfo{booktitle}{Advances in Neural Information Processing
  Systems}, \bibinfo{year}{2018}, pp. \bibinfo{pages}{7775--7784}.
\bibitem[{Adebayo et~al.(2018)Adebayo, Gilmer, Muelly, Goodfellow, Hardt, and
  Kim}]{adebayo2018sanity}
\bibinfo{author}{J.~Adebayo}, \bibinfo{author}{J.~Gilmer},
  \bibinfo{author}{M.~Muelly}, \bibinfo{author}{I.~Goodfellow},
  \bibinfo{author}{M.~Hardt}, \bibinfo{author}{B.~Kim},
\newblock \bibinfo{title}{Sanity checks for saliency maps},
\newblock in: \bibinfo{booktitle}{Advances in Neural Information Processing
  Systems}, \bibinfo{year}{2018}, pp. \bibinfo{pages}{9505--9515}.
\bibitem[{Dimanov et~al.(2020)Dimanov, Bhatt, Jamnik, and
  Weller}]{dimanov2020you}
\bibinfo{author}{B.~Dimanov}, \bibinfo{author}{U.~Bhatt},
  \bibinfo{author}{M.~Jamnik}, \bibinfo{author}{A.~Weller},
\newblock \bibinfo{title}{You shouldn’t trust me: Learning models which
  conceal unfairness from multiple explanation methods},
\newblock in: \bibinfo{booktitle}{European Conference on Artificial
  Intelligence}, \bibinfo{year}{2020}.
\bibitem[{Poursabzi-Sangdeh et~al.(2018)Poursabzi-Sangdeh, Goldstein, Hofman,
  Vaughan, and Wallach}]{poursabzi2018manipulating}
\bibinfo{author}{F.~Poursabzi-Sangdeh}, \bibinfo{author}{D.~G. Goldstein},
  \bibinfo{author}{J.~M. Hofman}, \bibinfo{author}{J.~W. Vaughan},
  \bibinfo{author}{H.~Wallach},
\newblock \bibinfo{title}{Manipulating and measuring model interpretability},
\newblock \bibinfo{journal}{arXiv preprint arXiv:1802.07810}
  (\bibinfo{year}{2018}).
\bibitem[{Kim et~al.(2018)Kim, Wattenberg, Gilmer, Cai, Wexler, Vi{\'{e}}gas,
  and Sayres}]{tcav}
\bibinfo{author}{B.~Kim}, \bibinfo{author}{M.~Wattenberg},
  \bibinfo{author}{J.~Gilmer}, \bibinfo{author}{C.~J. Cai},
  \bibinfo{author}{J.~Wexler}, \bibinfo{author}{F.~B. Vi{\'{e}}gas},
  \bibinfo{author}{R.~Sayres},
\newblock \bibinfo{title}{Interpretability beyond feature attribution:
  Quantitative testing with concept activation vectors {(TCAV)}},
\newblock in: \bibinfo{editor}{J.~G. Dy}, \bibinfo{editor}{A.~Krause} (Eds.),
  \bibinfo{booktitle}{Proceedings of the 35th International Conference on
  Machine Learning, {ICML} 2018, Stockholmsm{\"{a}}ssan, Stockholm, Sweden,
  July 10-15, 2018}, volume~\bibinfo{volume}{80} of
  \textit{\bibinfo{series}{Proceedings of Machine Learning Research}},
  \bibinfo{publisher}{{PMLR}}, \bibinfo{year}{2018}, pp.
  \bibinfo{pages}{2673--2682}. \URLprefix
  \url{http://proceedings.mlr.press/v80/kim18d.html}.
\bibitem[{Zhou et~al.(2018)Zhou, Sun, Bau, and
  Torralba}]{zhou2018interpretable}
\bibinfo{author}{B.~Zhou}, \bibinfo{author}{Y.~Sun}, \bibinfo{author}{D.~Bau},
  \bibinfo{author}{A.~Torralba},
\newblock \bibinfo{title}{Interpretable basis decomposition for visual
  explanation},
\newblock in: \bibinfo{booktitle}{Proceedings of the European Conference on
  Computer Vision (ECCV)}, \bibinfo{year}{2018}, pp. \bibinfo{pages}{119--134}.
\bibitem[{Ghorbani et~al.(2019)Ghorbani, Wexler, Zou, and
  Kim}]{ghorbani2019towards}
\bibinfo{author}{A.~Ghorbani}, \bibinfo{author}{J.~Wexler},
  \bibinfo{author}{J.~Y. Zou}, \bibinfo{author}{B.~Kim},
\newblock \bibinfo{title}{Towards automatic concept-based explanations},
\newblock in: \bibinfo{booktitle}{Advances in Neural Information Processing
  Systems}, \bibinfo{year}{2019}.
\bibitem[{Yeh et~al.(2019)Yeh, Kim, Arik, Li, Ravikumar, and
  Pfister}]{yeh2019concept}
\bibinfo{author}{C.-K. Yeh}, \bibinfo{author}{B.~Kim}, \bibinfo{author}{S.~O.
  Arik}, \bibinfo{author}{C.-L. Li}, \bibinfo{author}{P.~Ravikumar},
  \bibinfo{author}{T.~Pfister},
\newblock \bibinfo{title}{On concept-based explanations in deep neural
  networks},
\newblock \bibinfo{journal}{arXiv preprint arXiv:1910.07969}
  (\bibinfo{year}{2019}).
\bibitem[{Kim et~al.(2017)Kim, Wattenberg, Gilmer, Cai, Wexler, Viegas, and
  Sayres}]{kim2017interpretability}
\bibinfo{author}{B.~Kim}, \bibinfo{author}{M.~Wattenberg},
  \bibinfo{author}{J.~Gilmer}, \bibinfo{author}{C.~Cai},
  \bibinfo{author}{J.~Wexler}, \bibinfo{author}{F.~Viegas},
  \bibinfo{author}{R.~Sayres},
\newblock \bibinfo{title}{Interpretability beyond feature attribution:
  Quantitative testing with concept activation vectors (tcav)},
\newblock \bibinfo{journal}{arXiv preprint arXiv:1711.11279}
  (\bibinfo{year}{2017}).
\bibitem[{Goyal et~al.(2019)Goyal, Shalit, and Kim}]{goyal2019explaining}
\bibinfo{author}{Y.~Goyal}, \bibinfo{author}{U.~Shalit},
  \bibinfo{author}{B.~Kim},
\newblock \bibinfo{title}{Explaining classifiers with causal concept effect
  (cace)},
\newblock \bibinfo{journal}{arXiv preprint arXiv:1907.07165}
  (\bibinfo{year}{2019}).
\bibitem[{Hinton(2007)}]{hinton2007learning}
\bibinfo{author}{G.~E. Hinton},
\newblock \bibinfo{title}{Learning multiple layers of representation},
\newblock \bibinfo{journal}{Trends in cognitive sciences} \bibinfo{volume}{11}
  (\bibinfo{year}{2007}) \bibinfo{pages}{428--434}.
\bibitem[{Zhou et~al.(2014)Zhou, Khosla, Lapedriza, Oliva, and
  Torralba}]{zhou2014object}
\bibinfo{author}{B.~Zhou}, \bibinfo{author}{A.~Khosla},
  \bibinfo{author}{A.~Lapedriza}, \bibinfo{author}{A.~Oliva},
  \bibinfo{author}{A.~Torralba},
\newblock \bibinfo{title}{Object detectors emerge in deep scene cnns},
\newblock \bibinfo{journal}{arXiv preprint arXiv:1412.6856}
  (\bibinfo{year}{2014}).
\bibitem[{Koh et~al.(2020)Koh, Nguyen, Tang, Mussmann, Pierson, Kim, and
  Liang}]{koh2020concept}
\bibinfo{author}{P.~W. Koh}, \bibinfo{author}{T.~Nguyen},
  \bibinfo{author}{Y.~S. Tang}, \bibinfo{author}{S.~Mussmann},
  \bibinfo{author}{E.~Pierson}, \bibinfo{author}{B.~Kim},
  \bibinfo{author}{P.~Liang},
\newblock \bibinfo{title}{Concept bottleneck models},
\newblock in: \bibinfo{booktitle}{Proceedings of Machine Learning and Systems
  2020}, \bibinfo{publisher}{International Conference on Machine Learning},
  \bibinfo{year}{2020}, pp. \bibinfo{pages}{11313--11323}.
\bibitem[{Isaac~Lage(2020)}]{LagDos20}
\bibinfo{author}{F.~D.-V. Isaac~Lage},
\newblock \bibinfo{title}{Human-in-the-loop learning of interpretable and
  intuitive representations},
\newblock in: \bibinfo{booktitle}{{ICML} Workshop on Human Interpretability},
  \bibinfo{year}{2020}. \URLprefix
  \url{http://whi2020.online/static/pdfs/paper_31.pdf}.
\bibitem[{Andrews et~al.(1995)Andrews, Diederich, and
  Tickle}]{andrews1995survey}
\bibinfo{author}{R.~Andrews}, \bibinfo{author}{J.~Diederich},
  \bibinfo{author}{A.~B. Tickle},
\newblock \bibinfo{title}{Survey and critique of techniques for extracting
  rules from trained artificial neural networks},
\newblock \bibinfo{journal}{Knowledge-based systems} \bibinfo{volume}{8}
  (\bibinfo{year}{1995}) \bibinfo{pages}{373--389}.
\bibitem[{Zilke et~al.(2016)Zilke, Menc{\'\i}a, and Janssen}]{zilke2016deepred}
\bibinfo{author}{J.~R. Zilke}, \bibinfo{author}{E.~L. Menc{\'\i}a},
  \bibinfo{author}{F.~Janssen},
\newblock \bibinfo{title}{Deepred--rule extraction from deep neural networks},
\newblock in: \bibinfo{booktitle}{International Conference on Discovery
  Science}, \bibinfo{organization}{Springer}, \bibinfo{year}{2016}, pp.
  \bibinfo{pages}{457--473}.
\bibitem[{Chen et~al.(2017)Chen, Fraiberger, Moakler, and
  Provost}]{chen2017enhancing}
\bibinfo{author}{D.~Chen}, \bibinfo{author}{S.~P. Fraiberger},
  \bibinfo{author}{R.~Moakler}, \bibinfo{author}{F.~Provost},
\newblock \bibinfo{title}{Enhancing transparency and control when drawing
  data-driven inferences about individuals},
\newblock \bibinfo{journal}{Big data} \bibinfo{volume}{5}
  (\bibinfo{year}{2017}) \bibinfo{pages}{197--212}.
\bibitem[{Krishnan et~al.(1999)Krishnan, Sivakumar, and
  Bhattacharya}]{krishnan1999extracting}
\bibinfo{author}{R.~Krishnan}, \bibinfo{author}{G.~Sivakumar},
  \bibinfo{author}{P.~Bhattacharya},
\newblock \bibinfo{title}{Extracting decision trees from trained neural
  networks},
\newblock \bibinfo{journal}{Pattern recognition} \bibinfo{volume}{32}
  (\bibinfo{year}{1999}).
\bibitem[{Sato and Tsukimoto(2001)}]{sato2001rule}
\bibinfo{author}{M.~Sato}, \bibinfo{author}{H.~Tsukimoto},
\newblock \bibinfo{title}{Rule extraction from neural networks via decision
  tree induction},
\newblock in: \bibinfo{booktitle}{IJCNN'01. International Joint Conference on
  Neural Networks. Proceedings (Cat. No. 01CH37222)},
  volume~\bibinfo{volume}{3}, \bibinfo{organization}{IEEE},
  \bibinfo{year}{2001}, pp. \bibinfo{pages}{1870--1875}.
\bibitem[{Kazhdan et~al.(2020)Kazhdan, Shams, and Li{\`o}}]{kazhdan2020marleme}
\bibinfo{author}{D.~Kazhdan}, \bibinfo{author}{Z.~Shams},
  \bibinfo{author}{P.~Li{\`o}},
\newblock \bibinfo{title}{Marleme: A multi-agent reinforcement learning model
  extraction library},
\newblock \bibinfo{journal}{arXiv preprint arXiv:2004.07928}
  (\bibinfo{year}{2020}).
\bibitem[{Liu et~al.(2008)Liu, Liao, and Carin}]{liu2008semi}
\bibinfo{author}{Q.~Liu}, \bibinfo{author}{X.~Liao},
  \bibinfo{author}{L.~Carin},
\newblock \bibinfo{title}{Semi-supervised multitask learning},
\newblock in: \bibinfo{booktitle}{Advances in Neural Information Processing
  Systems}, \bibinfo{year}{2008}.
\bibitem[{Matthey et~al.(2017)Matthey, Higgins, Hassabis, and
  Lerchner}]{dsprites17}
\bibinfo{author}{L.~Matthey}, \bibinfo{author}{I.~Higgins},
  \bibinfo{author}{D.~Hassabis}, \bibinfo{author}{A.~Lerchner},
  \bibinfo{title}{dsprites: Disentanglement testing sprites dataset},
  \bibinfo{howpublished}{https://github.com/deepmind/dsprites-dataset/},
  \bibinfo{year}{2017}.
\bibitem[{Wah et~al.(2011)Wah, Branson, Welinder, Perona, and
  Belongie}]{wah2011caltech}
\bibinfo{author}{C.~Wah}, \bibinfo{author}{S.~Branson},
  \bibinfo{author}{P.~Welinder}, \bibinfo{author}{P.~Perona},
  \bibinfo{author}{S.~Belongie},
\newblock \bibinfo{title}{The caltech-ucsd birds-200-2011 dataset}
  (\bibinfo{year}{2011}).
\bibitem[{LeCun et~al.(1990)LeCun, Boser, Denker, Henderson, Howard, Hubbard,
  and Jackel}]{lecun1990handwritten}
\bibinfo{author}{Y.~LeCun}, \bibinfo{author}{B.~E. Boser},
  \bibinfo{author}{J.~S. Denker}, \bibinfo{author}{D.~Henderson},
  \bibinfo{author}{R.~E. Howard}, \bibinfo{author}{W.~E. Hubbard},
  \bibinfo{author}{L.~D. Jackel},
\newblock \bibinfo{title}{Handwritten digit recognition with a back-propagation
  network},
\newblock in: \bibinfo{booktitle}{Advances in neural information processing
  systems}, \bibinfo{year}{1990}, pp. \bibinfo{pages}{396--404}.
\bibitem[{Srivastava et~al.(2014)Srivastava, Hinton, Krizhevsky, Sutskever, and
  Salakhutdinov}]{dropout}
\bibinfo{author}{N.~Srivastava}, \bibinfo{author}{G.~Hinton},
  \bibinfo{author}{A.~Krizhevsky}, \bibinfo{author}{I.~Sutskever},
  \bibinfo{author}{R.~Salakhutdinov},
\newblock \bibinfo{title}{Dropout: A simple way to prevent neural networks from
  overfitting},
\newblock \bibinfo{journal}{Journal of Machine Learning Research}
  \bibinfo{volume}{15} (\bibinfo{year}{2014}) \bibinfo{pages}{1929--1958}.
  \URLprefix \url{http://jmlr.org/papers/v15/srivastava14a.html}.
\bibitem[{Szegedy et~al.(2016)Szegedy, Vanhoucke, Ioffe, Shlens, and
  Wojna}]{szegedy2016rethinking}
\bibinfo{author}{C.~Szegedy}, \bibinfo{author}{V.~Vanhoucke},
  \bibinfo{author}{S.~Ioffe}, \bibinfo{author}{J.~Shlens},
  \bibinfo{author}{Z.~Wojna},
\newblock \bibinfo{title}{Rethinking the inception architecture for computer
  vision},
\newblock in: \bibinfo{booktitle}{Proceedings of the IEEE conference on
  computer vision and pattern recognition}, \bibinfo{year}{2016}, pp.
  \bibinfo{pages}{2818--2826}.
\bibitem[{Krizhevsky et~al.(2012)Krizhevsky, Sutskever, and
  Hinton}]{krizhevsky2012imagenet}
\bibinfo{author}{A.~Krizhevsky}, \bibinfo{author}{I.~Sutskever},
  \bibinfo{author}{G.~E. Hinton},
\newblock \bibinfo{title}{Imagenet classification with deep convolutional
  neural networks},
\newblock in: \bibinfo{booktitle}{Advances in neural information processing
  systems}, \bibinfo{year}{2012}, pp. \bibinfo{pages}{1097--1105}.
\bibitem[{Cui et~al.(2018)Cui, Song, Sun, Howard, and Belongie}]{cui2018large}
\bibinfo{author}{Y.~Cui}, \bibinfo{author}{Y.~Song}, \bibinfo{author}{C.~Sun},
  \bibinfo{author}{A.~Howard}, \bibinfo{author}{S.~Belongie},
\newblock \bibinfo{title}{Large scale fine-grained categorization and
  domain-specific transfer learning},
\newblock in: \bibinfo{booktitle}{Proceedings of the IEEE conference on
  computer vision and pattern recognition}, \bibinfo{year}{2018}, pp.
  \bibinfo{pages}{4109--4118}.
\bibitem[{Fong and Vedaldi(2018)}]{fong2018net2vec}
\bibinfo{author}{R.~Fong}, \bibinfo{author}{A.~Vedaldi},
\newblock \bibinfo{title}{Net2vec: Quantifying and explaining how concepts are
  encoded by filters in deep neural networks},
\newblock in: \bibinfo{booktitle}{Proceedings of the IEEE conference on
  computer vision and pattern recognition}, \bibinfo{year}{2018}, pp.
  \bibinfo{pages}{8730--8738}.
\bibitem[{Zhou et~al.(2004)Zhou, Bousquet, Lal, Weston, and
  Schölkopf}]{Zhou04learningwith}
\bibinfo{author}{D.~Zhou}, \bibinfo{author}{O.~Bousquet},
  \bibinfo{author}{T.~N. Lal}, \bibinfo{author}{J.~Weston},
  \bibinfo{author}{B.~Schölkopf},
\newblock \bibinfo{title}{Learning with local and global consistency},
\newblock in: \bibinfo{booktitle}{Advances in Neural Information Processing
  Systems 16}, \bibinfo{year}{2004}.
\bibitem[{Pedregosa et~al.(2011)Pedregosa, Varoquaux, Gramfort, Michel,
  Thirion, Grisel, Blondel, Prettenhofer, Weiss, Dubourg, Vanderplas, Passos,
  Cournapeau, Brucher, Perrot, and Duchesnay}]{scikit-learn}
\bibinfo{author}{F.~Pedregosa}, \bibinfo{author}{G.~Varoquaux},
  \bibinfo{author}{A.~Gramfort}, \bibinfo{author}{V.~Michel},
  \bibinfo{author}{B.~Thirion}, \bibinfo{author}{O.~Grisel},
  \bibinfo{author}{M.~Blondel}, \bibinfo{author}{P.~Prettenhofer},
  \bibinfo{author}{R.~Weiss}, \bibinfo{author}{V.~Dubourg},
  \bibinfo{author}{J.~Vanderplas}, \bibinfo{author}{A.~Passos},
  \bibinfo{author}{D.~Cournapeau}, \bibinfo{author}{M.~Brucher},
  \bibinfo{author}{M.~Perrot}, \bibinfo{author}{E.~Duchesnay},
\newblock \bibinfo{title}{Scikit-learn: Machine learning in {P}ython},
\newblock \bibinfo{journal}{Journal of Machine Learning Research}
  \bibinfo{volume}{12} (\bibinfo{year}{2011}).
\bibitem[{Maaten and Hinton(2008)}]{maaten2008visualizing}
\bibinfo{author}{L.~v.~d. Maaten}, \bibinfo{author}{G.~Hinton},
\newblock \bibinfo{title}{Visualizing data using t-sne},
\newblock \bibinfo{journal}{Journal of Machine Learning Research}
  \bibinfo{volume}{9} (\bibinfo{year}{2008}) \bibinfo{pages}{2579--2605}.

\end{thebibliography}

\appendix

\section{Concept Decomposition} \label{bottleneck_app}

The results and findings presented in existing work on concept-based explanations suggests that users often think of tasks in terms of concepts and concept interactions (see Section \ref{sec:concept-based-expl} for further details). For instance, consider the task of determining the species of a bird from an image. A user will typically perform this task by first identifying relevant concepts (e.g. wing color, head color, and beak length) present in a given image, and then using the values of these concepts to infer the bird species, in a bottom-up fashion.

On the other hand, Machine Learning (ML) models usually rely on high-dimensional data representations, and infer task labels directly from these high-dimensional inputs (e.g. a CNN produces a class label from raw input pixels of an image).

Consequently, \textit{Concept Decomposition} (CD) approaches attempt to explain the behaviour of such ML models by decomposing their processing into two distinct steps: concept extraction, and label prediction. In concept extraction, concept information is extracted from the high-dimensional input data. In label prediction, concept information is used to produce the output label. Hence, CD approaches attempt to explain ML model behaviour in terms of human-understandable concepts and their interactions in a bottom-up fashion, paralleling human-like reasoning more closely. 

Importantly, whilst this work focuses on CNN models and tasks, the notion of CD can in principle be applied to \textit{any} ML model and task.

\subsection{CBMs}

CBMs can be seen as a special case of models performing CD, in which CD behaviour is \textit{enforced by design}. Hence, these models explicitly consist of two submodels, with the first submodel extracting concept information, and the second submodel using this concept information for producing task labels. Importantly, non-CBM models can still demonstrate CD behaviour. For instance, the dSprites Task 2 model was shown to have CD behaviour, with relevant concept information extracted in the dense layers, and used for classification decisions.

\subsection{CBMs \& CME}

The utility of CBMs is that they produce models explicitly encouraged to use CD. Consequently, these models are much more likely to rely on the desired concepts during decision-making, and be more aligned with a user's mental model of the corresponding task. 

However, a given DNN model can already exhibit CD behaviour, and use the desired concept information (e.g. as was the case with both dSprites task models). In this case, costly modifications and model re-training are unnecessary. As discussed in Section \ref{methodology}, CME can extract concept information from pre-trained DNNs by training $L*k$ concept predictors (where $L$ denotes the number of DNN layers used in concept extraction, and $k$ denotes the number of concepts). As demonstrated in Section \ref{results}, these concept predictors can consist of simpler models (e.g. LRs), trained on only a fraction of the DNN training data. Thus, the computational cost of training these concept predictors is significantly smaller, compared to training a bottleneck model on all the training data, as done in the case of CBMs.

More importantly, CBM models require knowledge of existing concepts and available concept annotations. In practice, these annotations are often expensive to produce, especially for large datasets and/or a large number of concepts. Furthermore, information about which concepts are relevant and/or sufficient for solving a given task is often not fully available either. Instead, CME is capable of using existing DNN models to extract this information automatically in a semi-supervised fashion, making concept discovery (identifying the relevant concepts), and concept annotation both faster and cheaper. 

Overall, CME permits efficient interaction with pre-trained DNN models, which can be used to leverage concept-related knowledge stored in these models. Consequently, we believe that CME will be invaluable in situations where concept-related information is expensive/difficult to obtain, or is only partially-known.
In these cases, a user may interact with existing DNN models via CME, in order to refine existing concept-related knowledge.

It should be noted that a CBM can trivially be approximated using CME, by defining $\hat{p}$ as the output of a CBM's concept bottleneck layer, and defining $\hat{q}$ as the CBM's submodel producing task labels from the bottleneck layer output.

\subsection{Further Discussion}

As discussed in Section \ref{methodology}, CME explores whether a DNN is concept-decomposable, by attempting to approximate it with an extracted model that is concept-decomposable by design (i.e. explicitly consists of two separate stages). Intuitively, if a given DNN learns and relies on concept information of the specified concepts during label prediction, this concept information will be contained in the DNN latent space. Hence, the DNN decision process could be separated into two steps: concept information extraction, and consequent task label prediction.

Importantly, existing CD-based approaches (such as those discussed in Section \ref{rel_cbm}) require the set of concepts and their values to be (i) \textit{sufficient} to solve the corresponding classification task (i.e. the class labels can be predicted from concept information with high accuracy) (ii) learnable from the data (i.e. the DNN model will be able to learn concept information from the given dataset), in order to achieve high task performance.

However, these works do not discuss how to handle cases where these assumptions do not hold (e.g. as was the case with the CUB task). Thus, exploring ways of efficiently \textit{discovering} relevant concepts sufficient for solving a given task, as well as ways of ensuring whether this concept information is \textit{learnable} from the data are both important research directions for future work.

\section{dSprites Dataset} \label{dsprites_a}

\subsection{Description}

dSprites is a dataset of 2D shapes, procedurally generated from 6 ground truth independent concepts (color, shape, scale, rotation, x and y position). Table~\ref{dsprites_features} lists the concepts, and corresponding values. dSprites consists of $64 \times 64$ pixel black-and-white images, generated from all possible combinations of these concepts, for a total of $1 \times 3 \times 6 \times 40 \times 32 \times 32 = 737280$ total images.

\begin{table}[pos=htbp]
\caption{dSprites concepts and values}
\centering
\begin{tabular}{| c | c |} \hline

\textbf{Name}       & \textbf{Values}                               \\ \hline

Color               & white                                         \\ \hline
Shape               & square, ellipse, heart                        \\ \hline
Scale               & 6 values linearly spaced in $ [0.5, 1] $      \\ \hline
Rotation            & 40 values in $[0, 2 \pi]  $                    \\ \hline
Position X          & 32 values in $[0, 1]$                         \\ \hline
Position Y          & 32 values in $[0, 1]$                         \\ \hline

\end{tabular}
\label{dsprites_features}
\end{table}

\subsection{Pre-processing}

We select $16$ of the $32$ values for \textit{Position X} and \textit{Position Y} (keeping every other value only), and select $8$ of the $40$ values for \textit{Rotation} (retaining every 5th value). This step makes the dataset size more manageable (reducing it from $737280$ to $3*6*8*16*16=36864$ samples), whilst preserving its characteristics and properties, such as concept value ranges and diversity.

\section{Input-to-Concept Functions} \label{i_to_c_app}

\begin{figure*}[pos=t]
    \centering
    \includegraphics[scale=0.2]{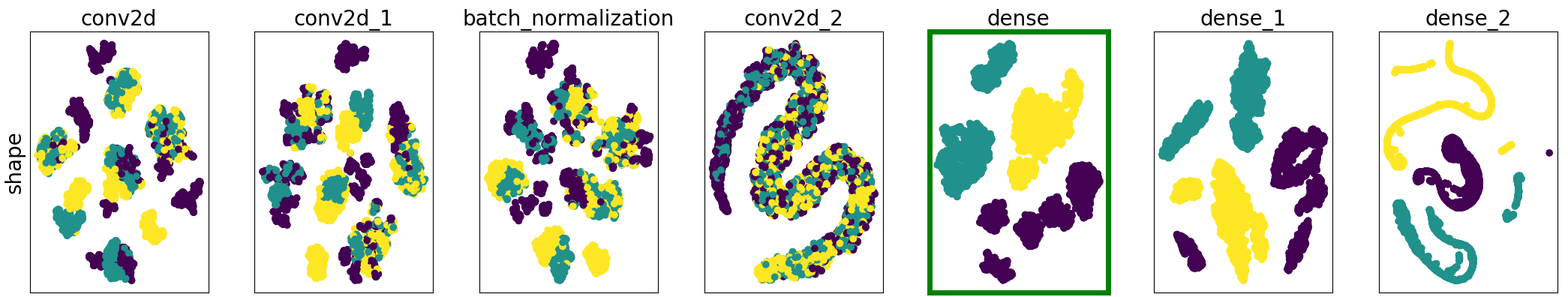}
    \caption{t-SNE plots for the relevant Task 1 concept. Each column corresponds to a different layer of the Task 1 model. Each plot is colored with respect to the concept's values. The subplot with a green border indicates the layer $\hat{p}$ uses for predicting the value of that concept}
    \label{fig:tsne_shape}
\end{figure*}

Figure~\ref{fig:tsne_shape} shows a t-SNE 2D projected plot of every layer's hidden space of the dSprites Task 1 model, highlighting different concept values of the relevant shape concept, and which layers were used by CME to predict it.

The CUB model has a considerably larger number of layers, and a considerably larger number of task concepts. Hence, for the sake of space, we demonstrate an example here using only 6 different model layers of the CUB model, and showing only the top $5$ important concepts identified in Section \ref{intervention_perf}. In this Figure, the concepts are named using their indices, and the layers are named following the naming convention used in \cite{koh2020concept}. Further details regarding layer naming and/or concept naming can be found in \footnote{https://github.com/yewsiang/ConceptBottleneck/tree/master/CUB}. For all concepts, concept values become significantly better-separated after the \verb|Mixed_7c| layer. However, the figure shows that concept values are still quite mixed together for some of the points, even for later layers. This low separability indicates that concept values will still be mis-predicted for some of the points, and that concept extraction for the CUB task will likely perform suboptimally.

\begin{figure*}[pos=t]
    \centering
    \includegraphics[scale=0.1]{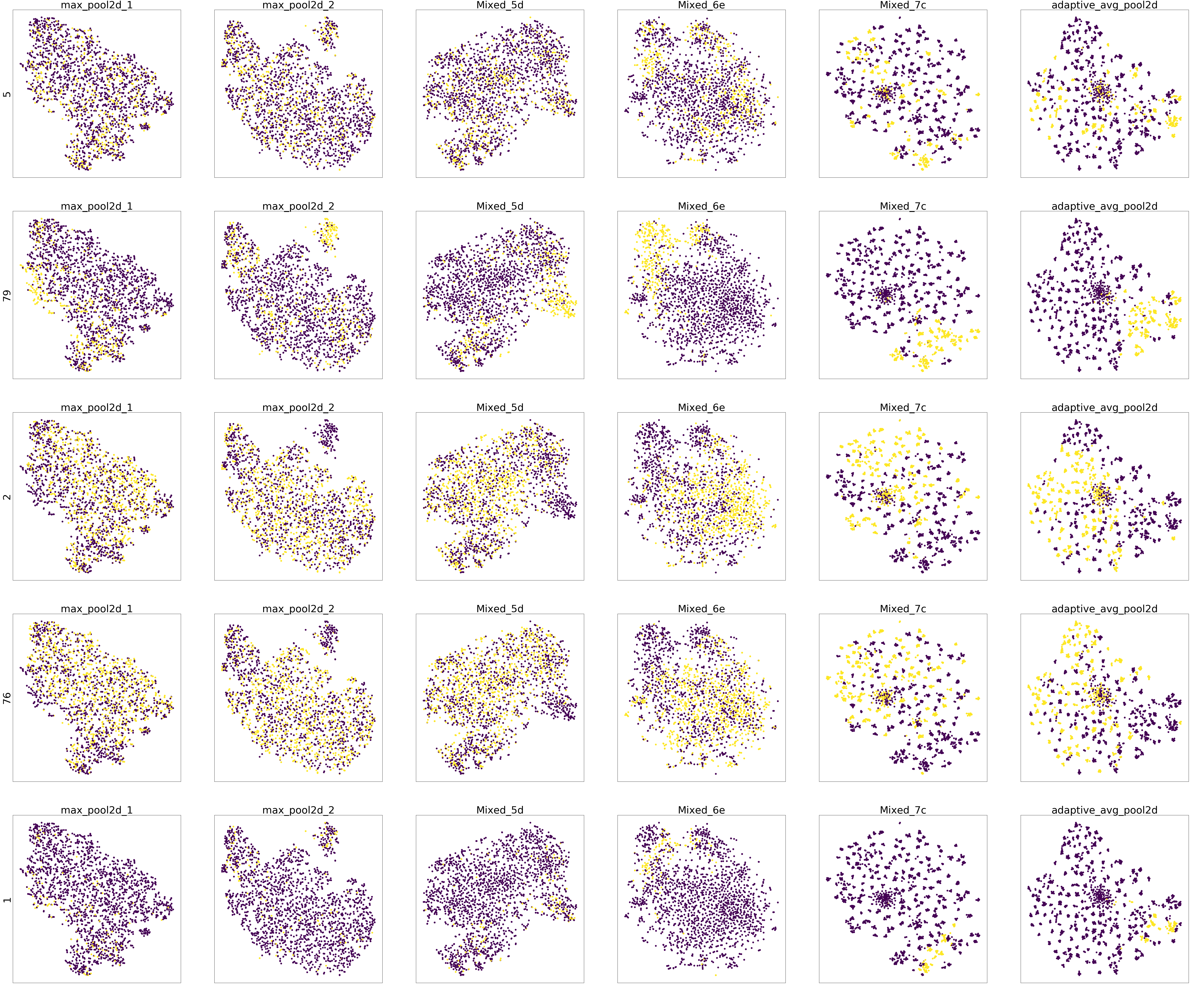}
    \caption{t-SNE plots for the top $5$ CUB concepts. Each column corresponds to a different layer of the CUB model. Each plot is colored with respect to the concept's values.}
    \label{fig:tsne_cub}
\end{figure*}

\section{Concept-to-Output Functions} \label{c_t_o_app}

\begin{figure*}[pos=t]
    \centering
    \includegraphics[scale=0.18]{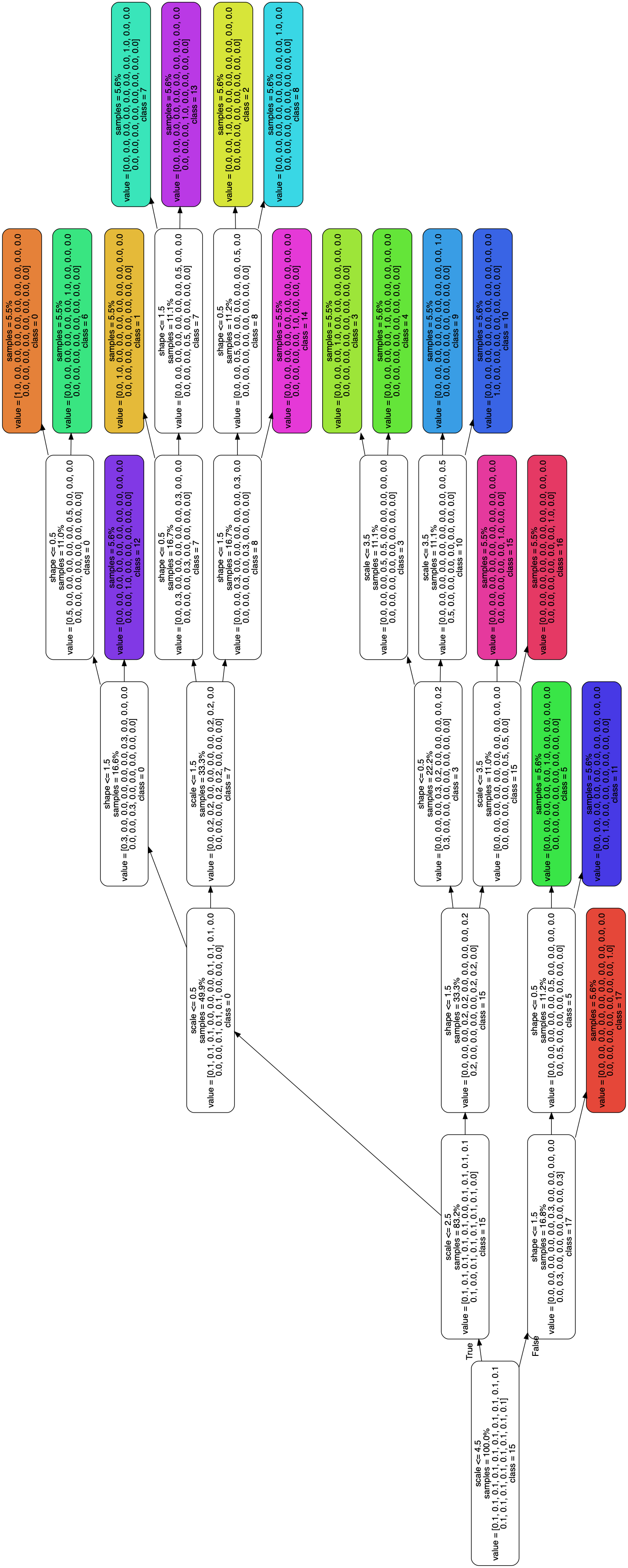}
    \caption{Visualisation of a decision tree $\hat{q}$ extracted from the Task 2 model. The model has correctly learned to differentiate between classes based on the shape and scale concept values.}
    \label{fig:output-to-concept-t2}
\end{figure*}

Figure \ref{fig:output-to-concept-t2} shows the decision tree extracted for dSprites Task 2.
Overall, this model has correctly learned to differentiate between classes based on the shape and scale concepts (note: there are $3\times 6$ shape and scale concept values, for a total of $18$ output classes).

\end{document}